\documentclass[final,5p,times,twocolumn]{elsarticle}

\usepackage[table]{xcolor}
\usepackage{colortbl}
\usepackage{balance}
\usepackage{amssymb}
\usepackage{array}
\usepackage{longtable}
\usepackage{booktabs}
\usepackage{multirow}
\usepackage{makecell}
\usepackage{threeparttable}
\usepackage{float}
\usepackage{graphicx}
\usepackage{amsthm}      
\usepackage{makecell, cellspace, caption}
\usepackage{subcaption}
\usepackage[shortcuts,acronym]{glossaries}
\usepackage{pgfplots}
\DeclareUnicodeCharacter{2212}{−}
\usepgfplotslibrary{groupplots,dateplot}
\usetikzlibrary{patterns,shapes.arrows}
\pgfplotsset{compat=newest}

\usepackage{booktabs}
\usepackage{tabularx}

\newacronym{cavs}{CAVs}{Connected and Autonomous Vehicles}
\newacronym{v2x}{V2X}{Vehicle to Everything}
\newacronym{cps}{CPS}{Cooperative Perception System}
\newacronym{veins}{Vehicle in Network Simulation}{Veins}
\newacronym{lidar}{LiDAR}{Light Detection and Ranging}
\newacronym{radar}{RaDAR}{Radio Detection and Ranging}
\newacronym{cpm}{CPM}{Cooperative Perception Message}
\newacronym{etsi}{ETSI}{European Telecommunications Standards Institute}
\newacronym{fov}{FoV}{ Field of View}
\newacronym{mec}{MEC}{Multi-access Edge Computing}
\newacronym{ran}{RAN}{Radio Access Network}
\newacronym{cn}{CN}{Core Network}
\newacronym{bs}{BS}{5G base station}
\newacronym{eq}{UE}{user equipment}

\journal{Computer Communications}

\begin{document}

\begin{frontmatter}



\title{Leveraging the Edge and Cloud for V2X-Based Real-Time\\Object Detection in Autonomous Driving}


\author{Faisal Hawlader, François Robinet and Raphaël Frank}

\affiliation{organization={Interdisciplinary Centre
for Security, Reliability, and Trust (SnT)},
            addressline={University of Luxembourg},
            postcode={L-1855 },
            country={Luxembourg}}

\begin{abstract}
Environmental perception is a key element of autonomous driving because the information received from the perception module influences core driving decisions.
An outstanding challenge in real-time perception for autonomous driving lies in finding the best trade-off between detection quality and latency. 
Major constraints on both computation and power have to be taken into account for real-time perception in autonomous vehicles. 
Larger object detection models tend to produce the best results, but are also slower at runtime.
Since the most accurate detectors cannot run in real-time locally, we investigate the possibility of offloading computation to edge and cloud platforms, which are less resource-constrained. 
We create a synthetic dataset to train object detection models and evaluate different offloading strategies.
Using real hardware and network simulations, we compare different trade-offs between prediction quality and end-to-end delay.
Since sending raw frames over the network implies additional transmission delays, we also explore the use of JPEG and H.265 compression at varying qualities and measure their impact on prediction metrics.
We show that models with adequate compression can be run in real-time on the cloud while outperforming local detection performance.
\end{abstract}
\begin{keyword}
V2X; Object detection; Latency optimization; Edge computing; Cloud computing;


\end{keyword}

\end{frontmatter}
\section{Introduction}
\label{sec:introduction}
One of the core challenges in autonomous driving is to reliably and accurately perceive the environment around the vehicle.
Perception is crucial to ensure safe driving because the information received from this task influences the core driving decision which determines how the vehicle should plan its path. 
However, perception requires processing a large amount of sensor data (\textit{e.g.} camera, LiDAR, radar) in real-time. At the same time, the hardware embedded in vehicles is constrained by both cost and power consumption. Running all detection tasks locally can therefore require sacrifices on perception quality, in favour of real-time operation. 

In this work, we focus on visual perception using a front-facing camera. The position and class of objects in the scene is needed to plan collision-free paths in autonomous driving and should be available in real-time. We follow the existing literature and aim to perform object detection at a rate of 20Hz \cite{abdel20205g}. We investigate different variants of the YOLOv5 \cite{yolov5} detection model, which offer different detection qualities at different inference times. As illustrated in Table \ref{Tab:accuracy}, larger models generally perform better. However, higher performance comes at the cost of increased computational requirements. Due to the limited computing resources and power available in the car, running larger models can prove difficult.

One alternative is to offload some computations where resources are available. Compute capabilities are less limited on \ac{mec} platforms \cite{hawlader2023vehicle}, and the best hardware is available in the cloud. Offloading some perception computation to the cloud can be appropriate in some situations. However, data offloading to \ac{mec} or cloud adds some additional transmission latency, which might not be acceptable for time-sensitive applications. With the promise of new 5G technologies supported by C-V2X \cite{9345798}, data offloading is an interesting option to complement local perception in autonomous driving.
In our experiments, we used the Simu5G \cite{nardini2020simu5g} network simulator, leveraging the capabilities of Cellular Vehicle-to-Everything (C-V2X) communication. It's worth noting that we specifically used release 16 of the standard.

In order to reduce transmission delays associated with streaming raw frames over the network, we explore the use of H.265, a video compression that exploits the temporal relationships between frames \cite{matsubara2021neural}, and JPEG, a widely used method for image compression \cite{deguerre2019fast}. 
We evaluate these at varying qualities and measure their impact on detection quality. 
We investigate the integration of H.265 and JPEG streaming with edge and cloud platforms for real-time perception tasks. By leveraging the computational capabilities of these platforms, we demonstrate how the proposed streaming solutions can be seamlessly integrated into the object detection pipeline, maximizing detection quality while minimizing latency.

In this work, we explore data offloading strategies for remote object detection on edge and cloud devices. We aim to evaluate these strategies on their detection quality, as well as their compliance with end-to-end latency requirements using a realistic model of the communication channel.
We are making the following contributions to the use of edge and cloud technology for real-time perception in autonomous vehicles: 
\begin{itemize}
    \item We create a synthetic dataset to train an object detection model and evaluate the proposed offloading strategies.
    \item We investigate the transfer of camera frames and their processing on edge or cloud platforms. Using real hardware and network simulations, we compare different trade-offs between prediction quality and end-to-end delay.
    \item  We demonstrate a comprehensive framework that integrates H.265 and JPEG streaming using edge and cloud platforms for C-V2X based real-time object detection in autonomous driving. This solution addresses the challenge of transmitting camera sensor data with minimal latency while maintaining high-quality object detection.
\end{itemize}
The rest of the paper is organised as follows. In Section \ref{sec:related_work}, we review the related literature.
In Section \ref{sec:methodology}, we describe the hardware used, our network simulation settings, and the training and evaluation of our object detectors.
Section \ref{sec:results} presents the experimental results and discusses the different offloading trade-offs. 
Finally, in Section \ref{sec:conclusion} we conclude this work with an outline of our contributions and a discussion of future work directions.
\section{Related work}
\label{sec:related_work}
\subsection{V2I communication}
According to studies \cite{huang2020mobile, islam2021survey}, autonomous vehicles typically leverage vehicle-to-infrastructure (V2I) communication technology that allows the vehicles to offload the task of sensor data processing to a dedicated server.
The server could be placed on the edge using the 5G MEC \cite{computingframework} or in a cloud with higher computational resources \cite{khan2022survey}. V2I communications are carried out using the upload / download path \cite{nardini2020simu5g}. The autonomous car offloads the task to an edge or cloud server that performs computations and sends the results back to the car. 
In this use case, the local device performs only mandatory pre-processing tasks, such as compression / encoding \cite{siriwardhana2021survey}.
\begin{table}[!h]
    \centering 
    \resizebox{0.65\linewidth}{!}{%
    \begin{tabular}{l c c c  c c}
            & \multicolumn{3}{c}{\textbf{Model Size}} \\
        \cmidrule(l){2-4}
            \textbf{ } & Small & Large & \makecell{Large\\(high-res)} \\ 
        \midrule 
            All (mAP) & 0.64 & 0.66 & 0.85 \\ 
            Pedestrian & 0.30 & 0.36 & 0.81 \\ 
            Traffic light & 0.80 & 0.82 & 0.86 \\ 
            Vehicle & 0.79 & 0.81 & 0.89 \\ 
        \bottomrule 
    \end{tabular}
    }
    \caption{Average Precision results for different model variants (AP@50). Model variants are detailed in Section \ref{sec:methodology_hardware}}
    \label{Tab:accuracy}
\end{table}
Offloading sensor data to the cloud through V2I adds additional transmission latency \cite{tsao2001enhanced}.
According to \cite{nardini2020simu5g}, this communication latency could be reduced using MEC, which requires a low transmission delay compared to the cloud.
However, edge devices are also subject to limited resources that could limit application needs.
\subsection{Object detection and impact of compression}

\textbf{Object detection:}
Pioneering work in object detection has used hand-crafted procedures to extract features from raw images, before using them as inputs to one or more object detectors. Popular examples include the Viola-Jones face detector based on Haar-like features \cite{viola_jones_face_detector}, or the Histogram-of-Gradient detector \cite{hog_detector}. Recent years have seen the emergence of two families of detectors based on deep neural networks: two-stage and single-stage models. Two-stage detectors first roughly identify regions that are likely to contain an object, before filtering and refining these object proposals with a trained model \cite{rcnn,fastrcnn,fasterrcnn}. Although two-stage models achieve impressive accuracy, their computational complexity makes real-time operation a challenge. To remedy this, the SSD \cite{ssd_detector} and YOLO \cite{yolo} family of models propose combining region proposal and refinement into a single operation. In YOLO, input frames are divided into cells and a set of bounding boxes are predicted for each cell. The YOLO detector can be trained end-to-end using a loss function that accounts for bounding box accuracy, objectness probabilities, and class assignments. Our work leverages YOLOv5 \cite{yolov5}, a refined variant of the original YOLO method.

\textbf{Impact of compression on detection:} To enable faster data transmission for cloud inference, we study the impact of H.265 (video) and JPEG (image) compression on detection performance, for varying qualities. Existing work has shown that JPEG compression negatively affects the performance of models trained on uncompressed frames \cite{mitigating_compression_defects,impact_of_quality}. In the case of object detection, this effect is particularly noticeable at lower compression qualities. Performances deteriorate rapidly for moderate to heavy compression \cite{mitigating_compression_defects}. However, H.265 has recently gained significant attention because it offers superior video compression compared to its predecessor, making it an attractive choice for sensor data streaming. 
Although H.265 offers improved compression efficiency, high-quality video streams may still require significant bandwidth for transmission. Bandwidth limitations in edge-to-cloud communication or V2X networks may impact real-time streaming performance and object detection quality. Minimizing latency while maintaining the benefits of H.265 or JPEG compression poses a challenge that needs to be addressed.

\subsection{Perception using C-V2X communication}
Leveraging the advancements of 5G technologies, particularly edge and cloud computing, there is potential for C-V2X technology to be increasingly adopted and experimented \cite{siriwardhana2021survey, abdel20205g, computingframework}.
However, widespread application of C-V2X remained mostly in the experimental phase, with larger-scale deployments expected to follow as the 5G infrastructure continues to evolve and mature.
C-V2X qualifies to support advanced applications \cite{yu2021edge, vanholder2016efficient}, such as collective perception \cite{kovacs2022integrating}. Processing of perception sensor data using the onboard vehicle computer might not always be an option due to the limited computational power \cite{khan2022survey}. However, collective perception using the V2X service allows cars to potentially offload sensory data to edge and cloud for resource-intensive computations \cite{li2018deep, islam2021survey}. Perception data processing in \ac{mec} or the cloud appears to be a viable option for autonomous driving, which has been the subject of many studies \cite{nardini2020simu5g, khan2022survey, siriwardhana2021survey, islam2021survey}. However, the perception of surrounding objects requires real-time detection, which demands rapid processing and low latency. This cannot be arbitrarily achieved, as it is heavily dependent on where the data is processed. In \cite{kovacs2022integrating, belogaev2020cost, featuresharing}, different sensor data offloading strategies have been presented, ranging from raw data offloading to partially or completely processed data offloading to save network resources. However, their approach emphasises data communication to reduce transmission overhead without evaluating the impact on perception quality. In principle, offloading raw sensor data is excellent for perception accuracy \cite{ye2020cooperative}, but could increase transmission costs \cite{vnc_paper}. However, offloading compressed data can save network resources, but it might degrade detection quality \cite{ren2018distributed}. To the best of our knowledge, the literature does not yet provide a study of the trade-off between detection quality and end-to-end processing.
\section{Methodology}
\label{sec:methodology}
This section describes our initial efforts toward estimating the end-to-end delay of real-time object detection.
Following existing work, our objective is to perform object detection at a rate of 20Hz without degrading the detection quality \cite{abdel20205g}.
In this context, the end-to-end delay heavily depends on where we perform the computation.
If the car chooses to offload the data to an edge/cloud, it must exchange the raw data, which V2X technologies may not support due to bandwidth constraints \cite{vnc_paper}.
\subsection{Motivation \& Hardware}\label{sec:methodology_hardware}
\begin{figure}[!h]
\centering
\includegraphics
[width=0.9\linewidth]{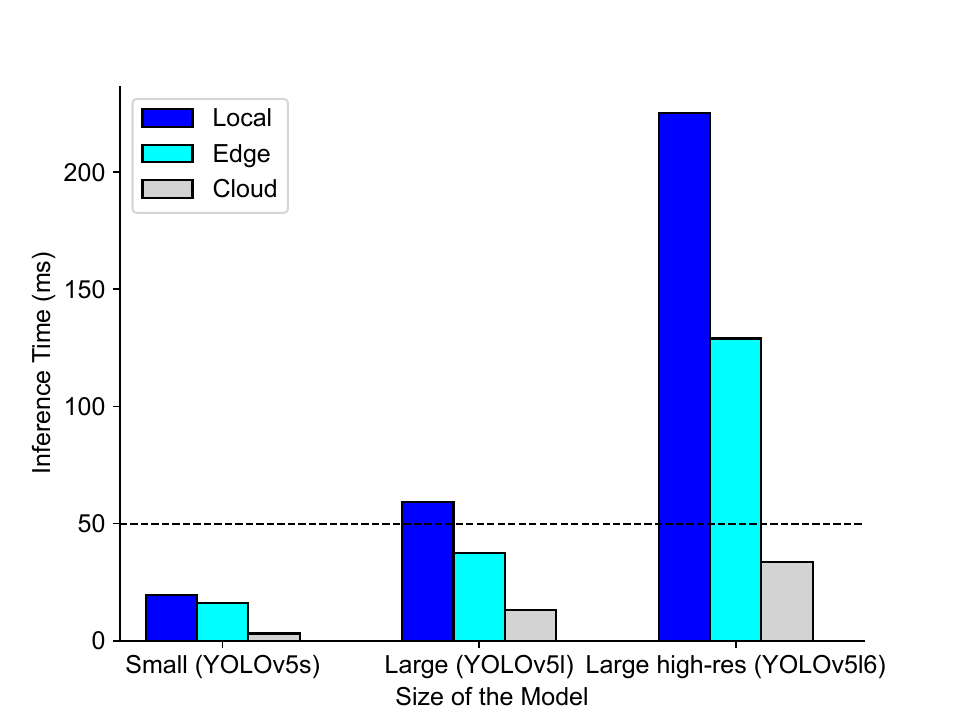}
\caption{Model inference time comparison between different model sizes on different platforms. We use half-precision floating-point computation at inference time in order to speed up computation.}
\label{fig:inference_time}
\end{figure}

The limited computing resources and energy consumption make it difficult to detect objects in the car.
Therefore, the computations are offloaded where the resources are available.
However, reaching high detection quality and decreasing the inference delay remain challenging \cite{ahmed2022survey}.
To better understand the problem and solve it, we performed a series of experiments on actual hardware setups.
The hardware configurations are shown in Table~\ref{Tab:Methogology_hardware}, where local represents the onboard device of an autonomous car that has limited processing power. However, the compute capabilities are less constrained on edge platforms, and the very best hardware is available in the cloud. This choice is made based on existing research \cite{ndikumana2022age, siriwardhana2021survey}.
\begin{table}[!h]
\begin{center}
\resizebox{0.98\linewidth}{!}{%
\begin{tabular}{lll}
\toprule
\textbf{Platform} & \textbf{Scenario / Model} &  \textbf{Hardware configuration}\\
\midrule
 Local & YOLOv5 small  & NVIDIA Jetson Xavier NX SoC\\ 
 ($\approx$20W) & 157 layers, 7M params & Volta GPU, 384 CUDA cores\\
& 640x640 Resolution & Carmel ARMv8.2 CPU@1.9GHz \\
\midrule
 Edge & YOLOv5 large & Laptop with GeForce GTX 1650\\ 
 ($\approx$100W) & 267 layers, 46M params & Turing GPU, 896 CUDA cores\\
& 640x640 Resolution & Intel i9-9980HK @2.4GHz\\
\midrule
Cloud & YOLOv5 large high-res & HPC node with Tesla V100\\ 
 ($\approx$450W) & 346 layers, 76M params  & Volta GPU, 5120 CUDA cores\\
& 1280x1280 Resolution & Intel Xeon G6132 @2.6 Ghz\\
\bottomrule
\end{tabular}}
\caption{Based on inference time constraints, we selected distinct platforms to run three models of different sizes. Figure~\ref{fig:inference_time} compares inference times for different models and platforms.}
\label{Tab:Methogology_hardware}
\end{center}
\end{table}
To perform object detection, we use YOLOv5 \cite{yolov5}, as it has been demonstrated to have superior performance in terms of accuracy and latency on state-of-the-art benchmark \cite{jocher2022ultralytics, zhao2022fast}.
YOLOv5 offers various model sizes ranging from small to large. We observe that larger models are beneficial for detection quality (see Table \ref{Tab:accuracy}), but there are inference timing constraints that must be taken into account. In this context, we aim to identify the best model for each platform that satisfies the time constraint.
We use Figure~\ref{fig:inference_time} to determine which model we can run on each platform. The inference time in Figure~\ref{fig:inference_time} shows that the larger model has an inference time of more than 50ms and as such is not suitable to run on the local platform. In view of the 20Hz detection speed, we decided to investigate the use of the small model on the local hardware, the large on the edge, and the large high-res one on the cloud. The exact YOLOv5 versions and the corresponding input resolutions are summarised in Table\ref{Tab:Methogology_hardware}.  

\subsection{Networking Aspects \label{sec:methodology_networking}}
\begin{figure}[!t]
\centering
\includegraphics[width=\linewidth]{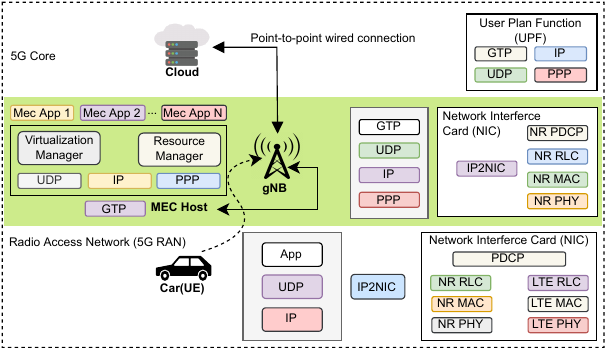}
\caption{Architecture of the end-to-end network simulation, showing the main elements of the 5G radio access network (RAN), including the multi-access edge computing (MEC) host-level components with a User Equipment (UE).}
\label{fig:networking_aspects}
\end{figure}

In this section, we describe the main elements of the 5G \ac{ran} and show how to use the network simulation framework to measure end-to-end network delays for a real-time object detection model supported by cloud infrastructure. To simulate the 5G data plane, we used Simu5G \cite{nardini2020simu5g} which is a OMNeT++ based discrete event network simulation library \cite{varga2010overview}. We focus on two scenarios, namely, perception data offloading with \ac{mec} and perception data offloading with cloud considering C-V2X.
The network environment we consider consists of a \ac{ran} and a 5G \ac{cn}.
The \ac{ran} has a single \ac{bs}, and one \ac{eq} is attached to the \ac{bs}, which is a vehicle in this use case \cite{farasat2021review}.
We also placed a \ac{mec} host close (500m) to the \ac{bs} connected to a wired network, so \ac{mec} can interact closely with \ac{bs} and obtain fast information from the \ac{ran} user.
The gNB is then connected to a cloud server via \ac{cn}.
The different components of the framework are shown in Figure~\ref{fig:networking_aspects}. 
We used a static setup in which the vehicle remained stationary throughout the experiments. 
We acknowledge, static setup is a limitation of our study, as it does not accurately reflect real-world scenarios where mobility is common.
However, we believe that it does not significantly undermine the validity of our findings, particularly in relation to network latency, which is relatively small.

\textbf{5G Core Network (CN):} we consider a standalone version of the 5G core network \cite{9211504}, which meets our requirements and is also available for simulation. 
A Point-to-Point (PPP) network interface is used to connect eNB to the cloud through a wired connection \cite{xu2021omnet}. The GPRS \cite{tsao2001enhanced} tunnelling protocol (GTP) is used to route IP datagrams (UDP) and establish a communication between gNB and the cloud.

\textbf{Multi-access Edge Computing (MEC):} The applications of MEC are growing and several standardisation initiatives are being carried out to provide a successful integration of MEC into the 5G network \cite{filali2020multi}. This work considers a simplified MEC host-level architecture in accordance with the \ac{etsi} reference \cite{computingframework}. In our case, an MEC host is co-located with gNB as shown in Figure~\ref{fig:networking_aspects}. The MEC host provides various modules that allow MEC applications to operate efficiently and seamlessly. The MEC applications run in a virtual environment and the resource manager orchestrates the life cycle of those applications. The Virtualisation Manager allocates, manages, and releases virtualised aids such as computing, storage, and networking resources.
The MEC host also includes a GTP protocol, which means that it can be located anywhere on the network. We placed the MEC 500m away from the gNB following the research is in \cite{nardini2020simu5g}. A PPP wired connection with 100G data rate is used to connect the MEC to gNB \cite{winick2002inet}.

\textbf{5G base station (gNB):} In the case of the scenario considered, gNB is configured with protocol up to Layer 3 and supports two network interface cards.
One for PPP wired connectivity to connect the core network and the other for the radio access network. The internal structure of the two network cards is shown in Figure~\ref{fig:networking_aspects}.
The PPP connection uses the GTP protocol, which has the same architecture as CN.
On the other hand, the radio access network card has four modules.
The topmost is the packet data convergence protocol (PDCP), which receives IP datagrams, performs ciphering, and sends them to the radio link control layer (RLC).
RLC service data units are stored in the RLC buffer and retrieved by the underlying Media Access Control layer (MAC) when a transmission is required.
The MAC layer aggregates the data into transport blocks, adds an MAC header, and sends everything through the physical layer (PHY) for transmission; for more details, we refer the reader to the Simu5G documentation \cite{nardini2020simu5g}.

\textbf{User equipment (UE):} As defined in the ETSI \cite{computingframework} and 3GPP specifications \cite{ali20213gpp}, user equipment is any device used by the end user. In our case, the user equipment refers to a car that is connected to the gNB, and equipped with C-V2X protocol stacks. We choose C-V2X because it is a 3GPP defined standard for connected mobility applications and works with 5G NR technology that is also available for simulation. It is important to mention that we used C-V2X only for bidirectional Vehicle-to-Infrastructure communications following the application needs. And the application we are focusing on is the offloading of perception data to the edge and cloud for real-time object detection. As part of the development policies and the implementation of Simu5G the UE has dual NIC to allow dual connectivity for both LTE \cite{khan2022survey} and 5G NR \cite{vanholder2016efficient}, as shown in Figure~\ref{fig:networking_aspects}.
\begin{table}[!h]
    \centering 
    \resizebox{1.00\linewidth}{!}{%
    \begin{tabular}{ll}
\toprule
\textbf{Parameter Name} & \textbf{Value}  \\ 
\midrule
Carrier frequency & 3.6 GHz \cite{frank2021poster} \\ 
gNB Tx Power & 46 dBm \\
Path loss model & \cite{nardini2020simu5g}, Urban Macro (UMa) \\
Fading + Shadowing model & Enable, Long-normal distribution \\
Number of repetitions & 200 \\
Path loss model & 3GPP-TR 36.873 \cite{rappaport2017overview}\\
UDP Packet size & 4096 B \cite{liu2020improve} \\
Throughput & 113.94 Mbit/s (Avg.) \\
Numerology $(\mu)$ & 3 \\
Latency (Vehicle-to-Edge) & 0.43 ms (Avg.) \\
Latency (Vehicle-to-Cloud) & 0.45 ms (Avg.) \\
Packet loss ratio & 0.0001 \\
\bottomrule
\end{tabular}}
\caption{Important network parameters with throughput and packet loss ratio for a stationary test. The averages are computed over 200 repetitions.}
    \label{Tab:network_parameters}
\end{table}

\subsection{Perception Aspects}\label{sec:methodology_perception}
\textbf{Dataset generation:} In order to train the YOLOv5 models described in Table \ref{Tab:Methogology_hardware}, we built a synthetic dataset using the CARLA simulator \cite{dosovitskiy2017carla}.
CARLA allows us to simulate a camera-equipped car driving in a rendered town environment, specifically under different weather conditions.
However, in this study, we only considered clear daylight weather conditions.
Additionally, it provides access to ground truth bounding boxes for three classes of interest: vehicles, pedestrians, and traffic lights.

We run the simulation to collect ten thousand camera frames, taken at 1Hz from the front of the ego-vehicle.
We also collect ground truth 3D bounding boxes, which we project into camera frame coordinates to obtain 2D bounding boxes usable for training and evaluating YOLOv5.
The dataset is split into three subsets: 6000 frames are used for training, 2000 for validation, and the remaining 2000 frames for testing.
Table~\ref{Tab:dataSet_instances} presents the distribution of instances in different classes and provides essential information on the composition of the data set, where the first column denotes partitioning for different purposes.
A notable aspect of this distribution is the considerably lower number of pedestrian instances compared to traffic lights and vehicles across all subsets.
This imbalance may influence the performance of models trained on the dataset, potentially making them less proficient in detecting pedestrians compared to traffic lights and vehicles.
Further analysis or data collection may be necessary to ensure representative and balanced data across all classes for more reliable results.
\begin{table}[!t]
    \centering 
    \resizebox{0.8\linewidth}{!}{%
    \begin{tabular}{l c c c  c c}
            & \multicolumn{3}{c}{\textbf{Class}} \\
        \cmidrule(l){2-4}
        & \textbf{Pedestrians} & \textbf{Traffic Lights} & \textbf{Vehicles} \\
        \midrule 
        Train & 12916& 43418& 33351 \\ 
        Validation & 2164 & 8272 & 11295 \\
        Test & 1756 & 11115 & 7897 \\  
        \midrule
        Total & 16836 & 62805 & 52576\\
        \bottomrule 
    \end{tabular}
    }
\caption{Dataset composition and instance counts per class.}
\label{Tab:dataSet_instances}
\end{table}
\subsection{Exploring Compression Settings}
\label{sec:compression_strategies}
We explore the influence of varying compression qualities on the volume of data to be transmitted to the edge or cloud.
Our in depth analysis of various compression levels aims to identify the most resource-efficient settings that can potentially optimize network resources and reduce the transmission latency, while preserving superior detection quality.

\begin{figure}[!h]
\centering
\includegraphics[width=0.85\linewidth]{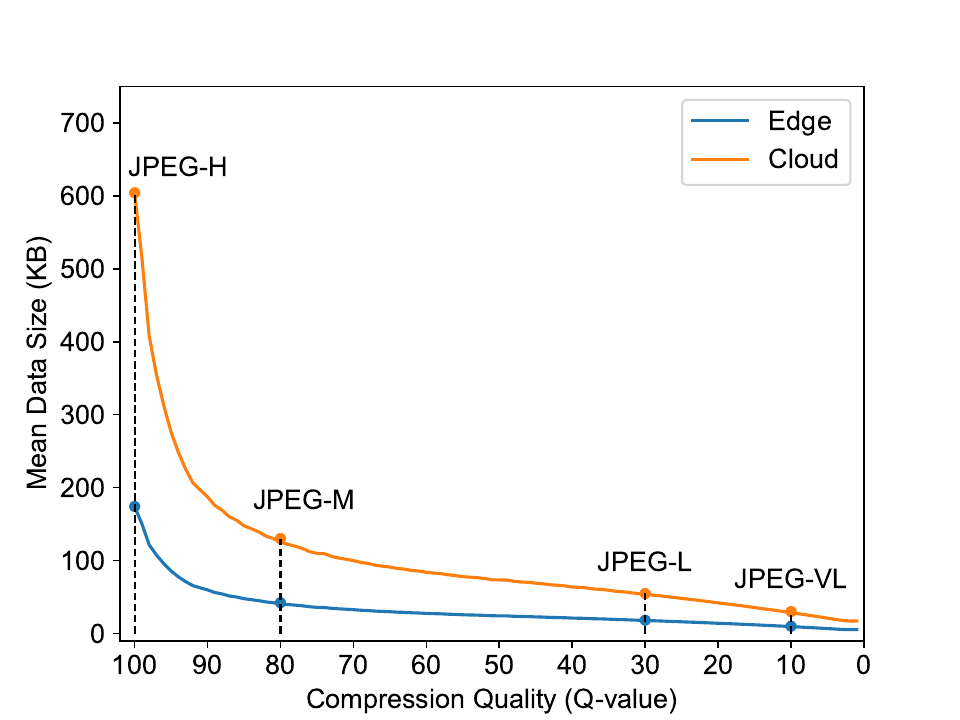}
\caption{The figure illustrates the mean data sizes for various JPEG compression qualities, namely JPEG-H (high quality), JPEG-M (medium quality), JPEG-L (low quality), and JPEG-VL (very low quality). We considered two distinct image dimensions: 640x640 for the edge and 1280x1280 for the cloud.}
    \label{Fig:dataSize_jpeg}
\end{figure}

\textbf{JPEG compression} is a widely used compression algorithm known for its ability to strike a balance between image quality and file size reduction.
It uses a lossy compression algorithm that selectively discards visual information to decrease file size.
The compression level is largely controlled by a quality parameter, the \textit{Q-value}, which ranges from 0 to 100.
Higher \textit{Q-values} preserve more details but result in larger sizes.
In this study, we evaluated JPEG compression across a spectrum of quality levels from high quality (JPEG-H) to very low quality (JPEG-VL), while maintaining the default values for other parameters.
The results, displayed in Figure~\ref{Fig:dataSize_jpeg}, demonstrate the mean data sizes associated with various JPEG qualities.
For further investigation, we defined four distinct scenarios: JPEG-H (high quality) at \textit{Q-value} 100, JPEG-M (medium quality) at 80, JPEG-L (low quality) at 30, and JPEG-VL (very low quality) at 10.

\textbf{H.265 compression} when using H.265 compression for camera sensor data streaming, several configuration and parameter settings can be adjusted.
The selection of these parameters plays a pivotal role in navigating the balance between compression quality and latency.
These settings include, but are not limited to, parameters such as Constant Rate Factor (CRF), presets, and lookahead settings, each offering various degrees of control over the encoding trade-off.
The following discussion will elaborate on these parameters and their impact on the overall outcome.
\begin{itemize}
\item  \textbf{H.265 Frames:} The fundamental computation of H.265 encoding starts with the estimation of three different frame types: I-frames (intra-coded frames), P-frames (predictive frames), and B-frames (bidirectional predictive frames).
Each of these uniquely contributes to overall efficiency.
The I-frame (known as the key frame) serves as the foundation, with all subsequent P-frame and B-frame relying on it.
The principle of I-frames is based on the fact that neighboring pixels within an image often exhibit high similarity.
Minor differences between these adjacent pixels can be encoded using fewer bits, reducing the overall size.

The P-frames differ from the I-frames in that they are not self-contained. Instead, the P-frames contain only the changes in the stream from the previous frame.
More specifically, a P-frame uses the prior I-frame or P-frame to encode the current frame, and hence making them predictive.
They look at what has changed (such as the movement of objects) since the previous frame.
If nothing has changed, no data need to be streamed, which is where much of the compression is achieved. 
If parts of the frame have changed, only the changes need to be encoded and streamed over the network.

However, B-frames refer to both the previous and future frames to achieve higher compression efficiency. 
They increase the encoding complexity and introduce some latency, but result in smaller file sizes and better quality. When H.265 is configured without B-frames, only I-frames and P-frames are used, reducing complexity and latency.

\item  \textbf{Lookahead} allows the encoder to examine future frames before encoding the current frame. This can increase compression efficiency and stream data quality, but can also introduce latency. If no lookahead is applied, this means that the encoder is not examining future frames before encoding the current frame. For the use case presented here, B-frames cannot be created since waiting for future information to encode the current frame implies an unacceptable increase of latency. Instead, stream compression may only rely on I-frames and P-frames. This setup is less efficient in terms of data compression compared to when B-frames are used, but it reduces latency and is suitable for real-time object detection in autonomous driving.

\item \textbf{Constant Rate Factor (CRF):} strives to maintain a steady visual quality taking into account the complexity and motion within each frame.
The H.265 compression assigns CRF values ranging from 0 to 51, which serve as quality-controlled variables affecting the bitrate.
For instance, the CRF value can be adjusted to control the trade-off between quality and data size that needs to be transmitted to the edge/cloud. A lower CRF value gives higher quality but a larger data stream size. Conversely, a higher CRF value provides a smaller frame size but lower quality.
\end{itemize}

\begin{figure}[!h]
    \centering
    \includegraphics[width=0.85\linewidth]{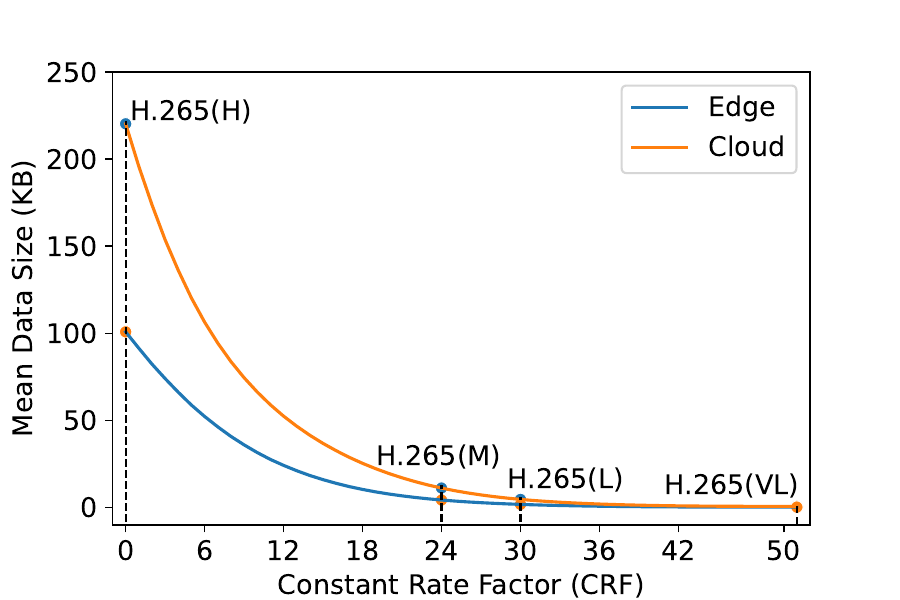}
    \caption{Mean data sizes for various H.265 compression factor, namely H.265-H (high quality), H.265-M (medium quality), H.265-L (low quality), and H.265-VL (very low quality). We considered two distinct image dimensions: 640x640 for the edge scenario and 1280x1280 for the cloud scenario.}
    \label{Fig:CRF-vs-dataSize}
\end{figure}

Using all possible CRF values, we performed a trade-off analysis to identify the optimal CRF for our real-time object detection use case. 
The lookahead functionality remained deactivated, and B-frames were excluded, while the rest of the parameters were set to their default values using FFmpeg \cite{ffmpeg}.
In the absence of B-frames and lookahead in the encoding setup, a notable reduction in both computational complexity and latency was observed.
These findings are particularly beneficial for applications that require real-time stream data transmission, where lower latency is a critical requirement.
The results of these experiments, depicted in Figure~\ref{Fig:CRF-vs-dataSize}, illustrate the relationship between the CRF values and the resulting data size.
Lower CRF values, which are indicative of higher quality, are correlated with larger data sizes. In contrast, increasing the CRF value leads to a reduction in the size of the data.
This clearly demonstrates the trade-off between data size and the range of quality levels from high quality (H.265-H) to very low quality (H.265-VL).
These observations will inform our decision-making on the quality levels that will be investigated further.
For further investigation, we defined four distinct scenarios: H.265-H (high quality) at CRF 0, H.265-M (medium quality) at 24, H.265-L (low quality) at 30, and H.265-VL (very low quality) at 51. We considered two distinct image dimensions: 640x640 for the edge scenario and 1280x1280 for the cloud scenario.
To facilitate a comprehensive evaluation, we categorized our analysis into four specific scenarios, each representing a different quality level in H.265 compression.
We assigned CRF 0 to H.265-H for high quality, CRF 24 to H.265-M for medium, CRF 30 to H.265-L for low, and finally, CRF 51 to H.265-VL for very low quality, as demonstrated in Figure~\ref{Fig:CRF-vs-dataSize}.

\textbf{Training Protocol:} All models are trained for 100 epochs on a single NVIDIA Tesla V100. We use the Adam optimizer with an initial learning rate of $0.001$. In order to speed up training, we set the batch size to the maximal value that fits in GPU memory for each experiment.
\section{Results}\label{sec:results}
This section describes the experiments carried out to evaluate our proposed data offloading strategies. Experiments are performed utilizing the hardware setup described in Section \ref{Tab:Methogology_hardware}.
Our analysis focuses on two critical key performance indicators, which are end-to-end delay and quality of detection. The close monitoring of these metrics enables the identification of potential bottlenecks or inefficiencies in real-time object detection for autonomous driving, ensuring that the vehicle responds accurately and in a timely manner to its surroundings.
\subsection{End-to-end delay evaluation}
To be able to compare the performance/latency trade-off of the three scenarios described in Section \ref{sec:methodology_hardware} and Table \ref{Tab:Methogology_hardware}, we first measure their end-to-end delays. 

In the case of the local platform, the delay depends only on the inference time on the local hardware.
Note that non-maximum suppression (NMS) \& input preprocessing are included in our inference time measurements, in addition to the forward pass of the model. By considering these integral components, we obtain a local end-to-end delay of $19.5$ ms.
This latency forms a critical part of our performance analysis, reflecting the efficiency of processing data locally without using any offloading strategies.

For the second and third scenarios, we evaluate the data offloading strategy between the car and edge or cloud platforms.
Network latency is measured using the end-to-end simulation framework presented in Figure \ref{fig:networking_aspects}. The important network simulation parameters with throughput and packet loss ratio are summarised in Table~\ref{Tab:network_parameters}. 
In these situations, the end-to-end delay includes compression, transmission, decompression and inference. 
Although the time required to send the results back is not directly indicated, we assume an additional latency of 0.43 ms for the transmission delay, predicated on the notion that the raw detection results can be accommodated within a single packet.
This assertion is supported by the data from Table~\ref{Tab:network_parameters}, which shows that a single packet of size 4096 B takes approximately 0.43 ms (Avg.) to travel from the edge/cloud to the vehicle.

Sending raw uncompressed frames to remote platforms results in large transmission delays because of the size of the data.
We measured an average end-to-end latency of $123.2$ ms in the vehicle-to-edge scenario. This delay rises to $521.7$ ms in the vehicle-to-cloud scenario due to the higher quality frame being processed by the cloud model.
Since these delays are not acceptable in most practical perception scenarios, we investigated the use of various compression strategies specified in Section~\ref{sec:compression_strategies} to reduce the volume of camera sensor data that needs to be transmitted to the edge or cloud over the network. Compression always occurs on the local device, and decompression happens on either the edge or cloud device, depending on the scenario.
As illustrated in Table~\ref{Tab:data_2Btransfer}, compression can drastically reduce the size of the frame to be transmitted, allowing real-time remote object detection when C-V2X is available. 

Table~\ref{Tab:data_2Btransfer} illustrates a comprehensive analysis of the data size and end-to-end delay for edge and cloud scenarios under varying JPEG and H.265 compression qualities. 
When examining the edge scenario, the data size without compression is 1.23 MB, causing an end-to-end delay of 123.20 ms.
The introduction of high quality JPEG compression significantly reduces the data size to 174.12 KB (\textit{i.e.} 13.82\% of the original), and the end-to-end delay accordingly decreases to 59.48 ms.
As the quality of JPEG compression is further reduced, we see a concomitant decrease in both data size and end-to-end delay, reaching as low as 9.48 KB (or 0.75\% of the original size) and a 37.27 ms delay with JPEG very Low compression.
We also observed significant improvements with H.265 compression. By exploiting the temporal relationships between frames, H.265 is able to minimize the data size to 0.26 KB (or 0.01\% of the original size), thus achieving a delay of 37.47 ms in a very low setting.

\begin{figure}[!t]
    \centering
    \subfloat[Car-to-Edge (large model, $640\times640$ input).]{
      \includegraphics[width=0.9\linewidth]{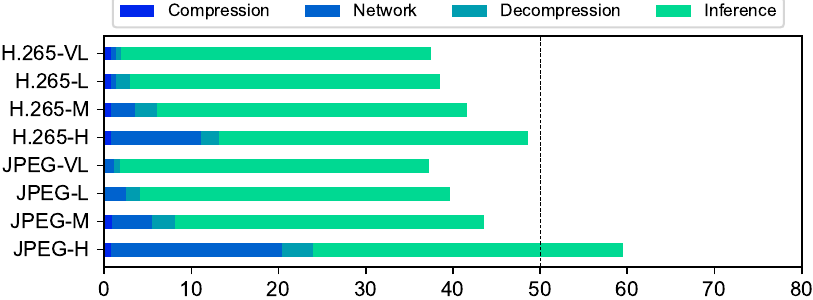}
      \label{fig:E2E-car2edge}} 
      \\
    \subfloat[Car-to-Cloud (large model, $1280\times1280$ input).]{
          \includegraphics[width=0.9\linewidth]{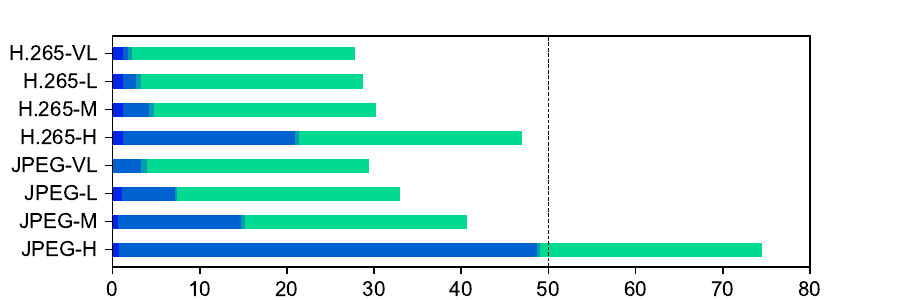}
          \label{fig:E2E-car2cloud}}
    \caption{Breakdown of average end-to-end delay (ms) into compression overhead, network transfer and inference times. The dashed line represents the latency constraint for real-time object detection in autonomous driving scenarios.}
   \label{fig:e2e-delays}
\end{figure}

In the cloud scenario with no compression, the data size is considerably larger, starting at 4.92 MB with an end-to-end delay of 521.7 ms.
However, similar to the edge scenario, the implementation of JPEG compression and H.265 compression shows a reduction in both the size of the data and the delay.
With very low JPEG compression, the data size is reduced to 28.60 KB (or 0.6\% of the original size) and results in a delay of 29.42 ms.
With H.265 compression at a very low setting, the data size decreases to 0.69 KB (or 0.01\% of the original) and the delay reduces to 27.78 ms. Table~\ref{Tab:data_2Btransfer} conclusively demonstrates that the use of compression techniques can significantly reduce data size and latency in both vehicle-to-edge and vehicle-to-cloud scenarios. A breakdown and further details of these end-to-end delays are provided in Figure~\ref{fig:e2e-delays}.

\begin{table}[!t]
\begin{center}
\resizebox{0.9\linewidth}{!}{%
\begin{tabular}{cllc}
\textbf{Platform} & \textbf{JPEG Quality} &  \textbf{\makecell{Avg. data size\\(\% of original)}} & \textbf{\makecell{End-to-end\\delay (ms)}}\\
\toprule
 & No compression & 1.23 MB (100\%) & 123.20\\ 
 \cmidrule(l){2-4}
 & JPEG-H & 174.12 KB (13.82\%) & 59.48 \\
  & JPEG-M & 40.78 KB (3.24\%) & 43.59 \\
 & JPEG-L & 17.86 KB (1.42\%) & 39.62\\
 \multirow{1}{*}{\begin{tabular}{c} Edge\\640$\times$640\end{tabular}}
& JPEG-VL & 9.48 KB (0.75\%) & 37.27\\
\cmidrule(l){2-4} 
& H.265-H & 100 KB (2.00\%)& 48.65\\
& H.265-M & 4.20 KB (0.09\%)& 41.61\\
& H.265-L &  1.80 KB (0.04\%)& 38.51\\
& H.265-VL & 0.26 KB (0.01\%)& 37.47\\
\midrule
 & No compression   & 4.92 MB (100\%) & 521.7\\ 
\cmidrule(l){2-4}
& JPEG-H & 604.38 KB (12.00\%) & 74.50 \\
 & JPEG-M & 125.51 KB (2.50\%) & 40.71 \\
& JPEG-L & 53.78 KB (1.13\%) & 32.93 \\
\multirow{1}{*}{\begin{tabular}{c} Cloud\\1280$\times$1280 \end{tabular}} 
& JPEG-VL & 28.60 KB (0.6\%) & 29.42\\
\cmidrule(l){2-4}
& H.265-H &  220 KB (4.37\%)& 46.93\\
& H.265-M & 11.20 KB (0.22\%)& 30.21\\
& H.265-L & 4.69 KB (0.09\%)& 28.72\\
& H.265-VL & 0.69 KB (0.01\%)& 27.78\\
\bottomrule
\end{tabular}}
\caption{Data size and end-to-end delay for both edge and cloud scenarios under different JPEG and H.265 compression qualities. The final column refers to the end-to-end delay in a vehicle-to-edge/cloud scenario. Figure~\ref{fig:e2e-delays}, illustrates a detailed breakdown of these delays.}
\label{Tab:data_2Btransfer}
\end{center}
\end{table}

A breakdown of average end-to-end delays for different compression qualities is shown in Figure~\ref{fig:e2e-delays}.
In all cases, compression and decompression have a negligible impact on the overall delay.
As expected, network transmission increases with the amount of data to be transmitted.
The inference times are constant for a given platform since the same model and input quality are considered.
In terms of end-to-end delay, all compression qualities are viable for real-time operation at 20Hz on both platforms, except the JPEG high quality.
The next section will investigate how compression impacts detection quality. 

Figure~\ref{fig:e2e-delays} provides a detailed breakdown of the average end-to-end delays for each of the various compression qualities evaluated.
In all cases, compression and decompression have a negligible impact on the overall delay.
As expected, network transmission latency increases with the volume of data to be streamed.
The inference times remain constant on a given platform, as the same model and input quality are consistently maintained.
From an end-to-end delay perspective, all of the assessed compression quality levels prove to be well-suited for real-time operation at 20Hz on both platforms, with the sole exception of JPEG-H high quality.
The next section examines how these compression strategies influence detection quality.
\subsection{Analyzing detection quality}
In order to obtain a complete picture of detection quality, measuring only Precision and Recall is insufficient.
We follow the object detection literature and compute the Average Precision (AP), which is the area below the Precision-Recall curve.
A detection is considered a true positive if its Intersection-over-Union (IoU) with a ground truth bounding box exceeds 50\%. In order to derive a single metric for all classes, the per-class APs are averaged to obtain the Mean Average Precision (mAP).

\begin{figure}[!h]
    \centering
    \includegraphics[width=0.85\linewidth]{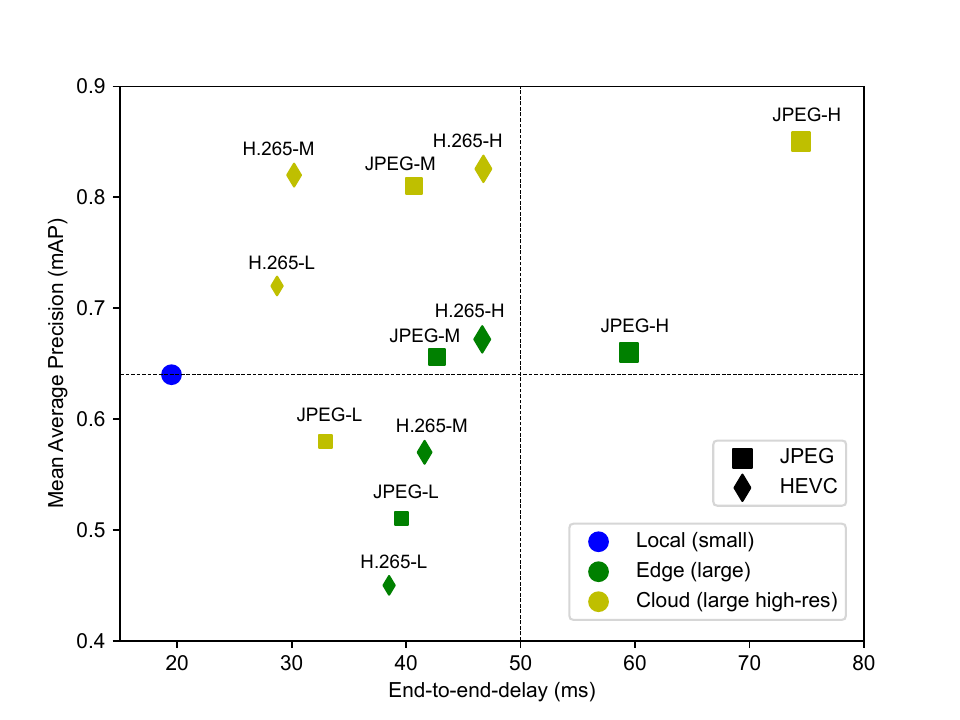}
    \caption{Trade-off between mean average precision (mAP) and end-to-end delay for different platforms and compression qualities. The end-to-end delay corresponds to the total of compression, network, decompression, and object detection inference delays. Very-low quality settings for both H.265-VL and JPEG-VL are not included, since their mAP fell below 10\%, an unacceptable detection rate regardless of the reduced end-to-end delays.}
    \label{Fig:E2E-vs-mAP-comcom}
\end{figure}

As already discussed, the end-to-end delay can be significantly reduced using various compression techniques. However, excessive compression affects the detection quality, degrading the mAP. This section aims to determine the best trade-offs between detection quality and end-to-end delay. These trade-offs are illustrated in Figure~\ref{Fig:E2E-vs-mAP-comcom}.
As expected, local operation is the fastest in terms of latency (19.5 ms), and obtains 64\% mAP. As demonstrated in Figure~\ref{Fig:E2E-vs-mAP-comcom}, the cloud platform consistently outperformed the edge platform in terms of mAP and end-to-end latency trade-off in all comparable compression techniques.
For example, on the cloud platform, when using H.265-H compression, the mAP reaches 82\%, but the end-to-end delay is 46.93 ms.
Interestingly, when using the H.265-M compression scenario, the mAP remains the same, but the delay is significantly reduced to 29.21 ms.
Furthermore, lowering the compression quality to H.265-L slightly decreases the mAP to 72\% and reduces the delay to 28.72 ms.
Figure~\ref{Fig:E2E-vs-mAP-comcom} also shows an increase in mAP from 58\% (JPEG-L) to 81\% (JPEG-M), reaching 85\% (JPEG-H) when considering JPEG compression techniques in the cloud.
At the same time, the delay increases from 32.93 ms (JPEG-L) to 40.71 ms (JPEG-M) and peaks at 74.50 ms (JPEG-H).
Although the highest detection quality (85\% mAP) is achieved with JPEG-H compression, this comes at the cost of an increased end-to-end latency of 74.5 ms, which is not suitable for real-time object detection at 20Hz. Nevertheless, it can still be used for applications where 10Hz is an acceptable latency.
\begin{figure*}[!t]
\centering
\begin{subfigure}[b]{0.48\textwidth}
\centering
\includegraphics[width=\textwidth]{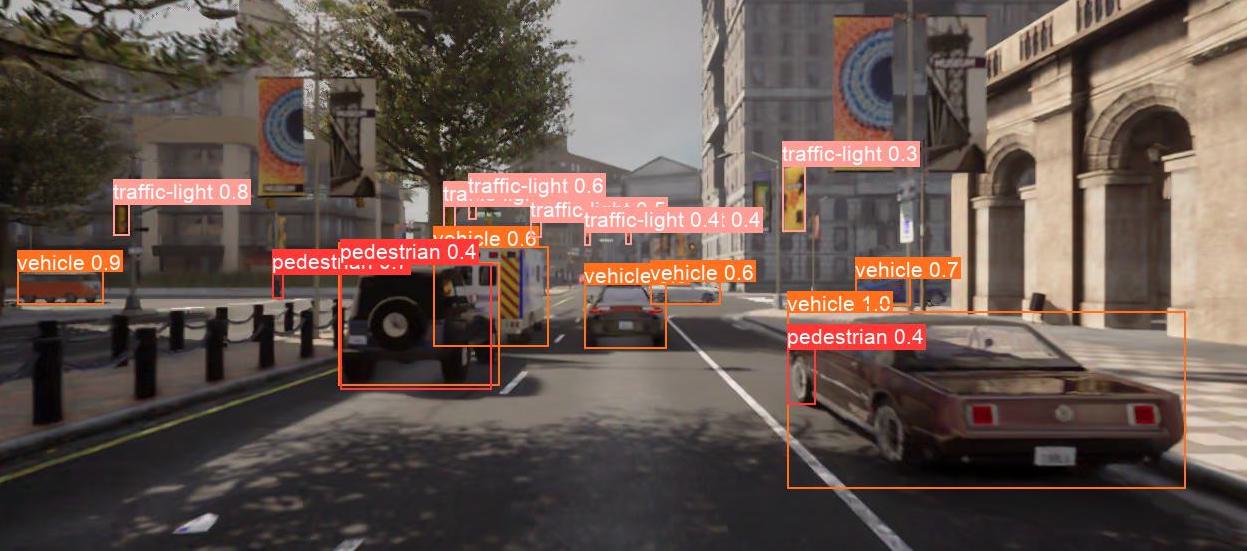}
\caption{H.265-H (CRF=0); Detected: 1 pedestrian, 7 vehicles, 8 traffic light}
\end{subfigure}
\begin{subfigure}[b]{0.48\textwidth}
\centering
\includegraphics[width=\textwidth]{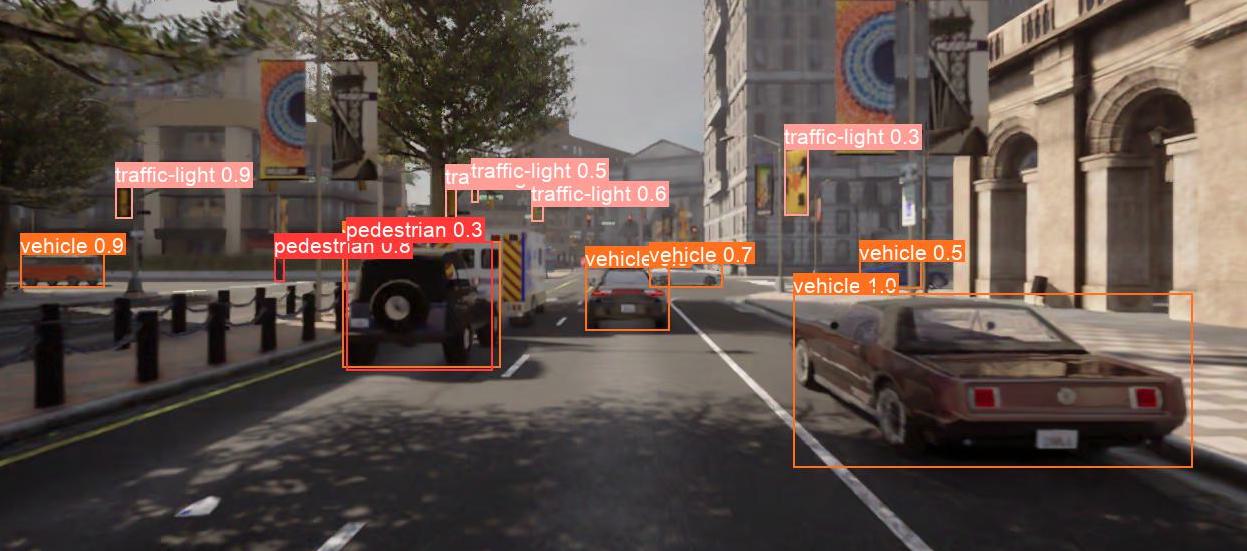}
\caption{H.265-M (CRF=24); Detected: 1 pedestrian, 7 vehicles, 7 traffic light}
\end{subfigure}
\begin{subfigure}[b]{0.48\textwidth}
\centering
\includegraphics[width=\textwidth]{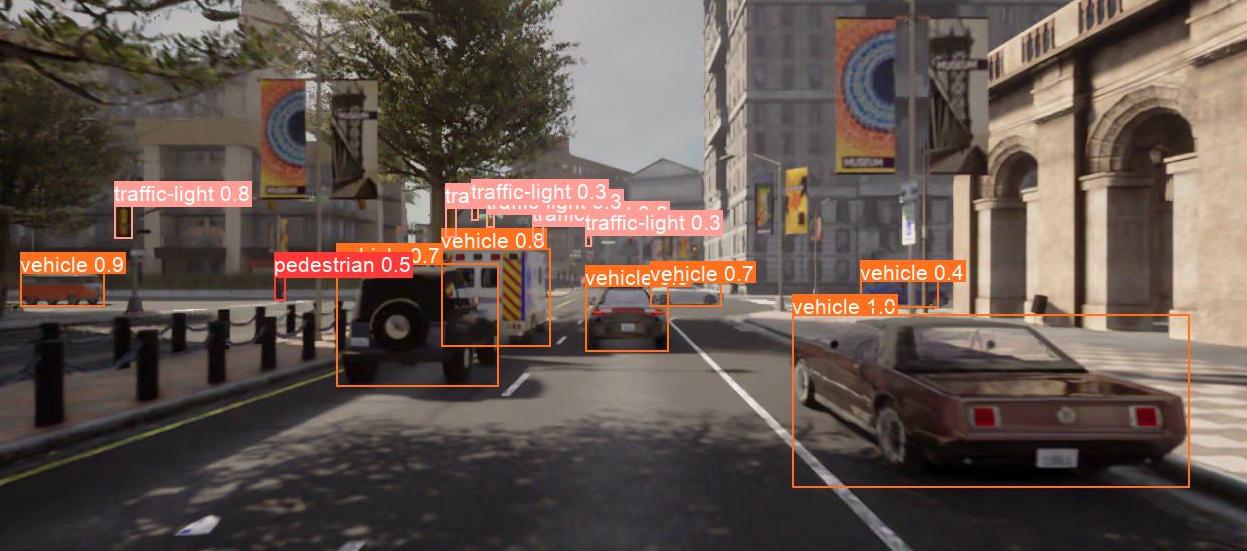}
\caption{H.265-L (CRF=30); Detected: 0 pedestrian, 7 vehicles, 6 traffic light}
\end{subfigure}
\begin{subfigure}[b]{0.48\textwidth}
\centering
\includegraphics[width=\textwidth]{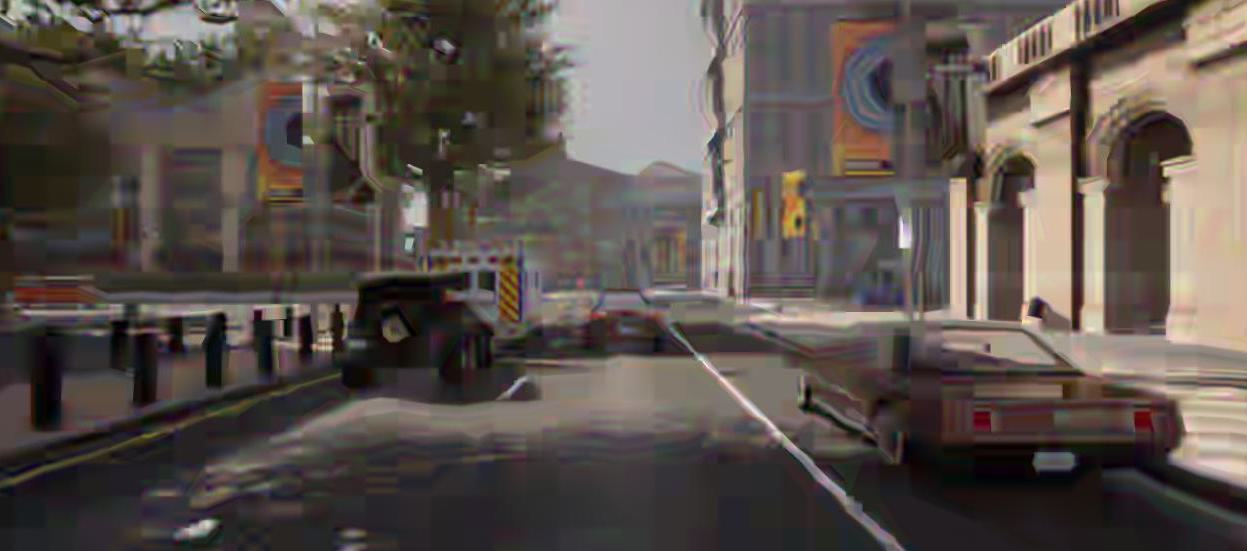}
\caption{H.265-VL (CRF=51); Detected: 0 pedestrian, 0 vehicles, 0 traffic light}
\end{subfigure}
\caption{Visualisation of the number of detected pedestrians, vehicles and traffic light in the cloud platform on different H.265 compression settings. A detection is considered only if its IoU with a ground truth bounding box exceeds 50\%. Ground truth: 8 traffic lights, 8 vehicles and 2 pedestrians.}
\label{fig:visulation_conditaions}
\end{figure*}
In the edge scenario, considering the H.265-High compression setting results in a mAP of 67\%, accompanied by an end-to-end delay of 48.65 ms.
For H.265-M and H.265-L, the mAP gradually respectively decreases to 57\% and 45\%, with corresponding delay reductions of 41.61 and 38.51 ms. In the JPEG case, mAP increases from 51\% (JPEG-L) to 66\% (JPEG-M) and then to 67\% (JPEG-H), while the delay increases from 39.62 ms (JPEG-L) to 43.59 ms (JPEG-M), reaching 59.48 ms (JPEG-H). 
Therefore, the edge scenario examined here is not advantageous over offloading to the cloud device. 
The use of better edge hardware specifically designed for mid-power inference rather than a traditional consumer GPU could most likely result in more competitive performance from the edge platform.
On the other hand, the cloud platform is interesting, as it offers better performance with both JPEG and H.265, with 81\% (JPEG-M) and 82\% (H.265-H) mAP, respectively.
Meanwhile, the end-to-end delays are kept under 50ms, respecting the 20Hz constraint.
Although compression is necessary for real-time operation on edge and cloud platforms, we observe its negative impact on detection quality. 
At extreme compression levels, remote detection mAP can drop below local performance while taking longer, rendering offloading harmful. For example, when using very low-quality settings for both H.265-VL and JPEG-VL in cloud or edge scenarios, mAP dropped below 10\%. This is considered unacceptable, regardless of the reduced end-to-end latency.
\begin{table}[!h]
    \centering 
     \resizebox{1.00\linewidth}{!}{%
    \begin{tabular}{ccccc}
        \textbf{\makecell{Compression}} & \textbf{Platform} &  \textbf{\makecell{Pedestrian\\(+\% of Local)}} & \textbf{\makecell{Vehicle\\(+\% of Local)}} & \textbf{\makecell{Traffic light\\(+\% of Local)}} \\
        \toprule
        \multirow{3}{*}{No compression} & \cellcolor{gray!40} Local & \cellcolor{gray!40} 0.30 (0\%)  & \cellcolor{gray!40} 0.79 (0\%)  & \cellcolor{gray!40} 0.80 (0\%) \\
                                            & Edge & 0.36 (+20\%) & 0.81 (+2\%) & 0.82 (+2\%)\\
                                            & Cloud & 0.81 (+170\%) & 0.89 (+12\%) & 0.86 (+7\%)\\
        \specialrule{1.4pt}{0pt}{0pt} 
        \multirow{2}{*}{JPEG-H} & Edge & 0.36 (+20\%) & 0.80 (+1\%) & 0.82 (+2\%) \\
                                     & Cloud & 0.80 (+166\%) & 0.89 (+12\%) & 0.86 (+7\%)\\
        \midrule
        \multirow{2}{*}{JPEG-M} & Edge & \textbf{0.41 (+36\%)} & \textbf{0.83 (+5\%)} & \textbf{0.78 (-2\%)} \\
                                     & Cloud & \textbf{0.65 (+116\%)} &\textbf{0.88 (+11\%)} & \textbf{0.85 (+6\%)}\\
        \midrule
        \multirow{2}{*}{JPEG-L} & Edge & 0.35 (+16\%) & 0.77 (-2\%)  & 0.74 (-7\%)\\
                                     & Cloud & 0.43 (+43\%) & 0.84 (+6\%) & 0.81 (+1\%)\\
        \midrule
        \multirow{2}{*}{JPEG-VL} & Edge & 0.23 (-23\%) & 0.73 (-7\%) & 0.55 (-31\%)\\
                                     & Cloud & 0.24 (-20.00\%) & 0.78 (-1\%) &  0.62 (-22\%)\\
        \specialrule{1.4pt}{0pt}{0pt} 
        \multirow{2}{*}{H.265-H} & Edge & \textbf{0.36 (+20\%)} & \textbf{0.80 (+1\%)} & \textbf{0.83 (+3\%)} \\
                                  &  Cloud & \textbf{0.68 (+126\%)} & \textbf{0.83 (+5\%)} & \textbf{0.83 (+3\%)} \\
        \midrule 
        \multirow{2}{*}{H.265-M} & Edge & 0.25 (-16\%) & 0.78 (-1\%) & 0.82 (+2\%) \\
                                  &  Cloud &  \textbf{0.69 (+130\%)} & \textbf{0.84 (+6\%)} & \textbf{ 0.82 (+2\%)} \\
        \midrule 
        \multirow{2}{*}{H.265-L} & Edge & 0.04 (-86\%) & 0.59 (-25\%) & 0.72 (-10\%) \\
                                  & Cloud & \textbf{0.51 (+70\%)} & \textbf{0.82 (+3\%)} & \textbf{0.82 (+2\%)} \\
        \midrule 
        \multirow{2}{*}{H.265-VL} & Edge & 0.01 (-96\%) &0.02 (-97\%) & 0.03 (-96\%)\\
                                  & Cloud & 0.03 (-90\%) & 0.23 (-70\%) & 0.04 (-95\%) \\
        \bottomrule 
    \end{tabular}}
    \caption{Per-class AP for different compression qualities and platforms. The input qualities are 640$\times$640 for the local and edge platforms, and 1280$\times$1280 for the cloud model. The baseline local model is highlighted in gray color, while all the models that are competitive, meeting both the mAP and latency constraint, are bolded.}
    \label{Tab:compression-with-accuracy}
\end{table}
We also evaluate the detection quality separately for the three classes of interest: pedestrians, vehicles, and traffic lights. A visual representation is shown in Figure~\ref{fig:visulation_conditaions}. The purpose of this evaluation is to understand the impact of compression on different classes. The results are shown in Table~\ref{Tab:compression-with-accuracy}. We observe that compression has a disproportionate impact on Average Precision (AP) for classes that are typically smaller in scale. 
We benchmark the performance of various compression settings with the corresponding platforms against the baseline scenario, \textit{i.e.} the local platform operating without any compression. 
The local platform performance shows an AP of 30\% for pedestrian detection and achieves an AP of 79\% and 80\% for vehicle and traffic light detection, respectively.
The Table~\ref{Tab:compression-with-accuracy} demonstrate that the use of a cloud platform, compared to local and edge platforms, led to significant performance improvements for pedestrian detection, particularly with medium compression settings for both JPEG-M (+116\%) and H.265-M (+130\%). 
In particular, high-quality compression (JPEG-H and H.265-H) enables both edge and cloud platforms to outperform the local scenario, with a cloud platform improving pedestrian detection by 166\% (JPEG-H) and 126\% (H.265-H) and an edge platform achieving a 20\% increase in AP in both cases.
Although all platforms perform similarly for vehicle and traffic light detection with high and medium compression settings, the performance decreases noticeably at lower compression settings. Under very low quality compression (JPEG-VL and H.265-VL), the AP for all three classes drops on both platforms, underlining the limitations of excessive compression. 
These findings underline the potential of edge and cloud platforms, paired with appropriate compression settings, to improve object detection performance relative to a local platform, especially for pedestrian detection. 
However, the benefits diminish with lower compression quality, emphasizing the need to strike the right balance between compression level and detection performance for real-time object detection in autonomous driving. 
\section {Conclusion and Future Work}
\label{sec:conclusion}
In this work, we have explored the possibility of real-time remote object detection. Although larger models perform better, they also require higher computational power.
Considering cost and power constraints in autonomous vehicles, the very best models cannot run locally in real-time.
To solve this problem, we have proposed different strategies to offload object detection to edge or cloud devices using C-V2X.
We have compared these strategies in terms of their detection quality and compliance with end-to-end latency requirements. 
To evaluate the proposed strategies, we have generated a synthetic dataset and have trained different variants of the YOLOv5 architecture.
Using an end-to-end 5G network simulation framework, we have measured the network latency incurred when transferring camera frames for processing on the edge and cloud.
We have also analysed how the use of heavy compression can reduce the frame size by up to 98\% when using JPEG and H.265 to enable real-time remote processing.
 We showed that models with adequate compression can be run in real-time on the edge/cloud while outperforming local detection performance. 

The experimental results demonstrated that the H.265 (video) compression technique generally offers better performance in terms of detection quality and end-to-end latency trade-off compared to JPEG (image), particularly in the cloud scenario.
However, there are scenarios where JPEG compression is still sufficient and can be used, such as where 10Hz is an acceptable latency.
Our experimental results show that excessive compression affects the detection quality compared to raw frames, particularly for the pedestrian class.

Future work will focus on testing the offloading strategies in different driving environments. Since local perception is still needed as a fallback to cope with bad connectivity, we plan on investigating the impact of mode switching between local and remote processing on detection quality and latency. 
Additionally, we will examine the influence of mobility on network latency, considering factors such as signal strength, handovers, obstacles, interference, and network congestion.
\section*{Acknowledgments}
This work is supported by the Fonds National de la Recherche of Luxembourg (FNR), under AFR grant agreement No 17020780 and project acronym \textit{ACDC}.
\bibliographystyle{elsarticle-num} 
\bibliography{globo_refs}



\end{document}